\documentclass[10pt,twocolumn,letterpaper]{article}

\usepackage{cvpr2026}              % To produce the CAMERA-READY version
\usepackage{amsmath,amsfonts}
\usepackage{algorithmic}
\usepackage{algorithm}
\usepackage{array}
\usepackage{textcomp}
\usepackage{verbatim}
\usepackage{booktabs}
\usepackage{bm}
\usepackage{multirow}
\usepackage{float}
\usepackage{amstext}
\usepackage{amsthm}
\usepackage{mathrsfs}
\usepackage{makecell}
\usepackage[table,xcdraw]{xcolor}
\usepackage{xcolor,colortbl}
\definecolor{best}{RGB}{255, 180, 180}     % 最优值浅红
\definecolor{second}{RGB}{255, 230, 180}   % 次优值浅橙
\definecolor{third}{RGB}{255, 255, 200}    % 第三名浅黄

\definecolor{cvprblue}{rgb}{0.21,0.49,0.74}
\usepackage[pagebackref,breaklinks,colorlinks,allcolors=cvprblue]{hyperref}

\def\paperID{3637} % *** Enter the Paper ID here
\def\confName{CVPR}
\def\confYear{2026}

\title{RaGS: Unleashing 3D Gaussian Splatting from 4D Radar and Monocular Cues \ for 3D Object Detection}

\author{
Xiaokai Bai$^{1}$,
Chenxu Zhou$^{2}$,
Lianqing Zheng$^{3}$,
Si-Yuan Cao$^{1}$,
Jianan Liu$^{4}$,\\
Xiaohan Zhang$^{1}$,
Yiming Li$^{1}$,
Zhengzhuang Zhang$^{5}$,
Hui-liang Shen$^{1}$\\[4pt]
$^{1}$College of Information Science and Electronic Engineering, Zhejiang University\\
$^{2}$College of Computer Science and Technology, Zhejiang University\\
$^{3}$School of Automotive Studies, Tongji University\\
$^{4}$Momoni AI, Gothenburg, Sweden\\
$^{5}$College of Energy Engineering, Zhejiang University\\[4pt]
{\tt\small shawnnnkb@gmail.com}
}

\begin{document}
\maketitle
\begin{abstract}
4D millimeter-wave radar is a promising sensing modality for autonomous driving, yet effective 3D object detection from 4D radar and monocular images remains challenging. Existing fusion approaches either rely on instance proposals lacking global context or dense BEV grids constrained by rigid structures, lacking a flexible and adaptive representation for diverse scenes. To address this, we propose RaGS, the first framework that leverages 3D Gaussian Splatting (GS) to fuse 4D radar and monocular cues for 3D object detection. 3D GS models the scene as a continuous field of Gaussians, enabling dynamic resource allocation to foreground objects while maintaining flexibility and efficiency. Moreover, the velocity dimension of 4D radar provides motion cues that help anchor and refine the spatial distribution of Gaussians. Specifically, RaGS adopts a cascaded pipeline to construct and progressively refine the Gaussian field. It begins with Frustum-based Localization Initiation (FLI), which unprojects foreground pixels to initialize coarse Gaussian centers. Then, Iterative Multimodal Aggregation (IMA) explicitly exploits image semantics and implicitly integrates 4D radar velocity geometry to refine the Gaussians within regions of interest. Finally, Multi-level Gaussian Fusion (MGF) renders the Gaussian field into hierarchical BEV features for 3D object detection. By dynamically focusing on sparse and informative regions, RaGS achieves object-centric precision and comprehensive scene perception. Extensive experiments on View-of-Delft, TJ4DRadSet, and OmniHD-Scenes demonstrate its robustness and SOTA performance. Code will be released.
\end{abstract}    
\section{Introduction}

\begin{figure}[ht]
    \centering
    \includegraphics[width=0.95\linewidth]{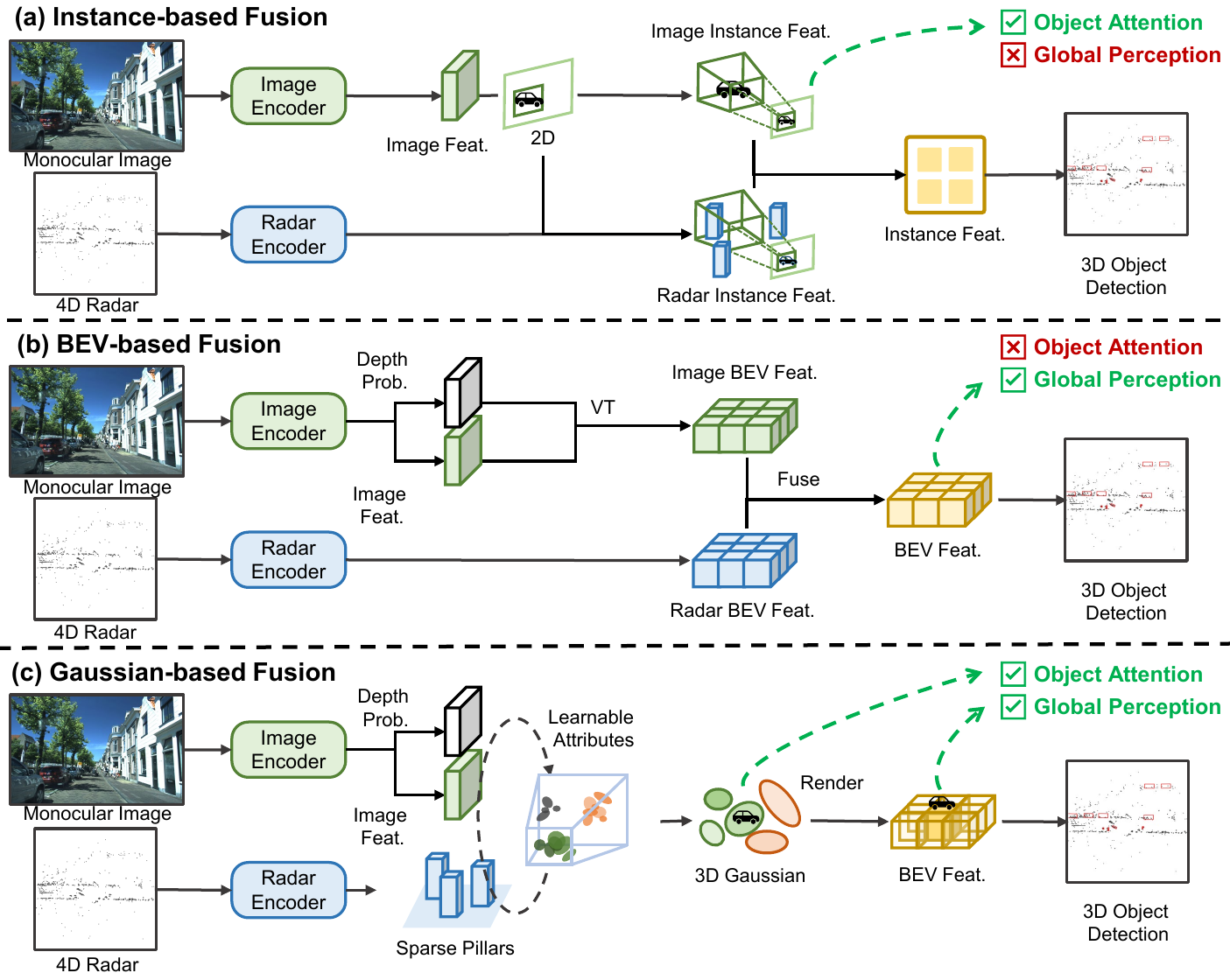}
    \caption{4D radar and camera fusion pipelines. (a) Instance-based fusion relies on 2D detection, limiting scene understanding. (b) BEV-based fusion uses predefined grids, causing inefficiencies in background modeling and fixed anchor sampling. (c) Our Gaussian-based fusion offers adaptive sparse objects attention while preserving scene perception.}
    \label{fig:comparison}
\end{figure}

Autonomous driving requires accurate 3D perception for safe decision-making \cite{Comprehensive_Survey}. Recently, 4D millimeter-wave radar has emerged as a highly promising sensor, known for its robustness in challenging environments, long-range detection capabilities, and ability to capture both velocity and elevation data. Cameras, on the other hand, provide high-resolution semantic information. The complementary nature of these two modalities is crucial for enhancing perception in autonomous driving, particularly for 3D object detecting. However, effective as well as efficient fusion across modalities remains a challenge, as the task is inherently sparse while still requiring a comprehensive understanding of the entire scene.

\begin{figure*}[htbp]
    \centering
    \includegraphics[width=0.95\linewidth]{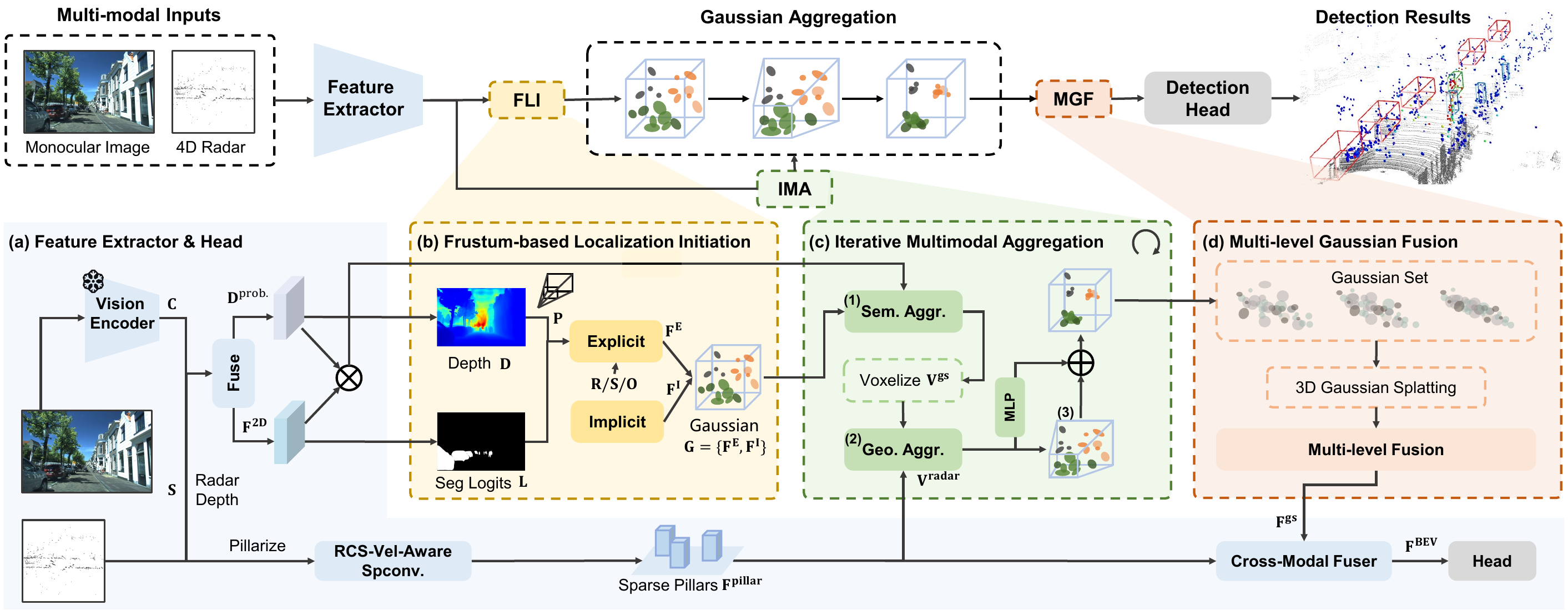}
    \caption{\textbf{Pipeline of RaGS.} RaGS consists of a Feature Extractor \& Head, Frustum-based Localization Initiation (FLI), Iterative Multimodal Aggregation (IMA), and Multi-level Gaussian Fusion (MGF). The positions of the Gaussians are initialized using the FLI module, along with learnable attributes such as rotation, scale, opacity, and implicit feature embeddings. These Gaussians are then passed into the IMA module, where they are projected onto the image plane to gather semantic information. Next, they are processed as voxels using sparse convolution with height-extended radar geometry, which implicitly utilizes radar velocity to guide residuals movement. Residuals relative to regions of interest are computed iteratively, updating the positions towards sparse objects. Finally, the multi-level Gaussians are rendered into Bird’s Eye View (BEV) features and fused through MGF, followed by cross-modal fusion for 3D object detection.}
    \label{fig:framework}
\end{figure*}

Recent research has focused on either instance-based or BEV-based fusion for 4D radar and camera 3D object detection. Instance-based approaches \cite{CRAFT, RADIANT, centerfusion} rely on 2D detectors to generate proposals, which are refined by aligning radar features. While these approaches directly maintain the 3D detection flow by focusing on sparse objects, they lack global scene understanding and are constrained by cascaded network designs, as shown in Fig. \ref{fig:comparison}(a). On the other hand, BEV-based approaches \cite{CRN, LXL, RCFusion, RCBEVDet, SGDet3D} project multimodal features into a predefined, fixed top-down space, enabling global reasoning. However, these approaches suffer from rigid voxelization, fixed anchor positions for sampling image semantics, and inefficiencies due to excessive background aggregation, as illustrated in Fig. \ref{fig:comparison}(b). While they offer promising performance in building scene perception, they are misaligned with the sparse nature of 3D object detection tasks, in which focusing primarily on foreground objects is significantly more efficient.

In parallel, 3D Gaussian Splatting (GS) \cite{3DGS} has emerged as a compact, continuous scene representation. Originally developed for neural rendering, GS models scenes as collections of anisotropic Gaussians optimized in 3D space, offering physical interpretability, sparsity, and flexibility. Recent extensions have applied GS to dynamic scenes \cite{DeformableGS} and LiDAR-camera fusion \cite{SplatAD}. However, these works have primarily focused on rendering or occupancy prediction \cite{Gaussianformer}, without exploring its potential for multi-modal fusion or 3D object detection, despite its natural alignment with the flexible needs of such tasks. Unlike 2D instance-based or BEV-based approaches, GS can dynamically allocate resources and adjust its attention, making it particularly efficient for identify interested objects within entire scene.

To address the limitations of existing fusion approaches and better adapt to the sparse nature of 3D object detection, we propose RaGS, a novel framework that leverages 3D Gaussian Splatting to fuse 4D radar and monocular cues, as presented in Fig. \ref{fig:comparison}(c). Unlike traditional dense voxel grids, RaGS represents the scene as a continuous field of 3D Gaussians, allowing for flexible aggregation without the constraints of fixed grid resolutions and fixed anchor sampling. By dynamically allocating resources iteratively, RaGS focuses on sparse foreground objects while preserving global scene perception through Gaussians’s continuous probability distribution. Specifically, we follow a cascaded pipeline composed of three key modules, as presented in Fig. \ref{fig:framework}. First, the Frustum-based Localization Initiation (FLI) identifies foreground regions and unprojecting pixels to initialize coarse, grounded 3D Gaussian localization. Next, the Iterative Multimodal Aggregation (IMA) aggregates radar and image features within frustum regions, refining Gaussian positions towards objects and enhancing both semantic and geometric expressiveness. Finally, the Multi-level Gaussian Fusion (MGF) integrates features rendered from Gaussians at different levels to produce a rich representation for object prediction. We evaluate RaGS on three 4D radar and camera benchmarks: View-of-Delft (VoD) \cite{VoD}, TJ4DRadSet \cite{TJ4D}, and OmniHD-Scenes \cite{OmniHD}, achieving state-of-the-art performance across the datasets, demonstrating the effectiveness of continuous Gaussian representation for multi-modal 3D object detection. Our contributions are summarized as
\begin{itemize}
\item We propose RaGS, the first framework that leverages 3D Gaussian Splatting to fuse 4D radar and monocular cam-era inputs for 3D object detection.
\item We design three modules to construct and refine the Gaussian from multi-modal inputs, enabling dynamic focus on objects while maintaining scene perception.
\item Extensive experiments on View-of-Delft, TJ4DRadSet and OmniHD-Scenes benchmarks demonstrate the effectiveness of our RaGS.
\end{itemize}

\section{Related Work}
\subsection{4D Radar based 3D Object Detection}
In recent years, research on 4D millimeter-wave radar has advanced rapidly, leading to the release of several benchmark datasets and an increasing focus on multi-modal fusion with cameras \cite{TJ4D, VoD, Kradar, OmniHD}.
Recent studies on 3D object detection using 4D radar–camera fusion typically follow two paradigms: instance-based and BEV-based paradigms. Instance-based approaches generate 2D proposals from images, refined with radar features. Examples include CenterFusion \cite{centerfusion} that aligns radar points with 2D boxes for better center predictions, CRAFT \cite{CRAFT} that integrates radar range and velocity cues, and RADIANT \cite{RADIANT} that recovers missed detections under occlusion. While effective for foreground localization, these methods depend heavily on proposal quality and lack global perception.

BEV-based approaches project multi-modal features into a top-down space for global spatial reasoning. RCFusion \cite{RCFusion} utilizes orthographic projection for semantics but neglects depth ambiguity, whereas other approaches like LXL \cite{LXL} and CRN \cite{CRN} introduce various improvements for image view transformation. L4DR \cite{L4DR} and Interfusion \cite{InterFusion}, on the other hand, further explored the role of LiDAR and 4D radar fusion for 3D object detection. Despite their strengths, BEV approaches rely on fixed-resolution voxel grids, leading to inefficiencies, excessive background aggregation, and reduced flexibility in 3D object detection. In contrast, we propose a dynamic 3D Gaussian-based representation, modeling multi-modal features as a continuous Gaussian field. This allows for more flexible and fine-grained fusion, making it particularly suitable for sparse 3D object detection.

\subsection{3D Gaussian Splatting in Autonomous Driving}
3D Gaussian Splatting (GS) has emerged as a key technique for real-time 3D scene reconstruction. Initially introduced as a faster alternative to neural radiance fields for rendering, GS has extended to tasks like non-rigid motion \cite{DeformableGS}, large-scale environments \cite{Drivinggaussian}, and multi-modal fusion with LiDAR-camera \cite{TCLC-GS, SplatAD} or radar-only inputs \cite{RadarSplat}.

Despite these advancements, current GS applications are predominantly in dense prediction tasks, such as rendering or occupancy prediction \cite{Gaussianformer}. However, its potential for 3D object detection remains largely unexplored, even though the sparse nature of detection task inherently aligns with Gaussian flexibility. While NeRF-based approaches \cite{Nerf-det, Mononerf} use volumetric representations for 3D detection, GS offers advantages like faster rendering and better integration with detection pipelines.

In this work, we leverage GS for sparse object modeling and scene perception. Unlike traditional BEV grid models that rely on fixed anchors for feature sampling, our method utilizes GS to enable dynamic feature aggregation by adaptively refining Gaussian positions towards targets, which perfectly aligns with the 3DGS foreground fitting insight pioneered by indoor models (3DGS-DET \cite{3DGS-DET}, Gaussian-Det \cite{Gaussian-Det}), although RaGS is designed for outdoor environments.

\setlength{\abovedisplayskip}{4pt}   % 公式上方间距
\setlength{\belowdisplayskip}{4pt}   % 公式下方间距
\setlength{\abovedisplayshortskip}{4pt}
\setlength{\belowdisplayshortskip}{4pt}
\section{Method}
% \subsection{Overview}
Fig. \ref{fig:framework} illustrates the overall architecture of our RaGS framework, which consists of four main components: Feature Extractor \& Head, Frustum-based Localization Initiation (FLI), Iterative Multimodal Aggregation (IMA), and Multi-level Gaussian Fusion (MGF). RaGS encodes image and radar depth into features, and radar point clouds into sparse geometric features. It then initializes 3D Gaussians via FLI, refines them with radar geometry and image semantics via IMA, and finally renders these Gaussians into multi-scale BEV features via MGF for 3D object detection.

% First, the feature extractor encodes the monocular image and radar depth into image features and a depth probability map, while the 4D radar point cloud is pillarized into sparse geometric features. Next, Gaussian aggregation is performed. Specifically, the FLI unprojects the foreground pixels from the perspective view into frustum space to initialize 3D Gaussians. These Gaussians are further refined and enriched via the IMA, which integrates radar geometry and image semantics while performing localization refinement. The MGF then renders the Gaussian set into multi-scale BEV-aligned features, which are finally fed into the multi-task head for depth estimation, semantic segmentation, and 3D object detection.

\subsection{Feature Extractor}
The feature extractor processes image and 4D radar. For the monocular image, we employ a ResNet-50 backbone with an FPN to extract and fuse multi-scale features $\mathbf{C} \in \mathbb{R}^{H \times W \times C}$, which are further fused with sparse radar depth $\mathbf{S} \in \mathbb{R}^{H \times W}$ to estimate depth probability $\mathbf{{D}^{\text{prob.}}} \in \mathbb{R}^{H \times W \times D}$ and enhanced feature map $\mathbf{F}^{\text{2D}} \in \mathbb{R}^{H \times W \times C}$. Here, $C$ and $D$ denotes the number of channel and depth bins, and $(H, W)$ denotes the resolution. The process can be expressed as $(\mathbf{F}^{\text{2D}}, \mathbf{{D}^{\text{prob.}}}) = \mathtt{Conv}(\mathtt{Concat(\mathbf{C}, \mathbf{S})})$, where $\mathtt{Conv}$ denotes convolution that output $(C+D)$ channel feature map.

Then, the metric depth $\mathbf{D}$ is computed by summing over predefined depth bins $\mathbf{D} = \sum_{d=1}^{D} P_d \cdot d$,
where \( P_d \) represents the probability of each depth bin \( d \). A lightweight segmentation network is further used to predict foreground regions from image features $\mathbf{F}^{\text{2D}}$, producing segmentation logits $\mathbf{L} \in \mathbb{R}^{H \times W}$. For the 4D radar point cloud, to fully utilize its RCS and velocity characteristics, we pillarize it and apply RCS-Vel-Aware sparse convolution to extract channel-specific information followed \cite{RCFusion}. These channel-splited features are then stacked into sparse pillars \( \mathbf{F}^{\text{pillar}} \in \mathbb{R}^{G \times C} \), where \( G \) is the number of pillars. This separate extraction helps avoid confusion and guides the subsequent feature aggregation of the Gaussian.

\subsection{Gaussian Aggregation}
\noindent\textbf{Frustum-based Localization Initiation (FLI).}
Gaussian initialization is crucial for guiding downstream feature aggregation. Previous approaches use learnable embeddings to initialize localization based on dataset-specific distributions. In this work, we initialize the position of each Gaussian using monocular cues, which provide a coarse but structured foreground prior focused on sparse objects. Thus, the Gaussians are modeled as a combination of two components: explicit physical properties (position \( \mathbf{P} \), rotation \( \mathbf{R} \), scale \( \mathbf{S} \), opacity \( \mathbf{O} \)) and implicit feature embeddings. The position is inferred from the FLI.

Specifically, we begin by selecting the top-K foreground pixels from the segmentation logits $\mathbf{L}$ and metric depth $\mathbf{D}$. Each selected foreground pixel \( (u, v) \) with its corresponding depth $d=\textbf{D}(u, v)$ is then unprojected into 3D space:
\begin{equation}
\mathbf{P}_{\text{unproj}} = d \cdot \mathbf{K}^{-1} (u, v, 1)^T,
\end{equation}
where \( \mathbf{K} \) is the camera intrinsic matrix. Furthermore, we also include 4D radar points \( \mathbf{P}_{\text{radar}} \) and randomly sample points in the frustum space to enrich spatial coverage and eliminate potential instability of foreground identification. For the random sampling process, we begin by defining the candidate points \(\mathbf{P}_{\text{cand}}\) as the centers of predefined voxels in the radar coordinate system. These points are projected onto the image space using the camera's extrinsic parameters: $(ud, vd, d) = \mathbf{K} \left( \mathbf{R} \mathbf{P}_{\text{cand}} + \mathbf{T} \right)$, where \( \mathbf{R} \) and \( \mathbf{T} \) are the rotation matrix and translation vector from radar to camera coordinates. We select points that lie within the image boundaries using the condition \( \mathbf{I} = \{ i \mid 0 \leq u_i < W, 0 \leq v_i < H \} \), and apply Furthest Sampling:
\begin{equation}
\mathbf{P}_{\text{sample}} = \mathtt{FurthestSampling}\left( \mathbf{P}_{\text{cand}}[\mathbf{I}] \right).
\end{equation}
The final position \( \mathbf{P} \) is a concatenation of the unprojected foreground points, 4D radar points, and sampled points:
\begin{equation}
\mathbf{P} = \mathtt{Concat}(\mathbf{P}_{\text{unproj}}, \mathbf{P}_{\text{sample}}, \mathbf{P}_{\text{radar}}) \in  \mathbb{R}^{N \times 3},
\end{equation}
where $N$ is the number of Gaussian anchors. This ensures that all initialized Gaussians are within the field of view (FoV) and offer coarse-but-structured initial localization, unlike BEV-based approaches that perceive the entire scene. The fully learnable explicit properties include a rotation vector \( \mathbf{R} \in \mathbb{R}^{N \times 4} \), scale \( \mathbf{S} \in \mathbb{R}^{N \times 3} \), and opacity \( \mathbf{O} \in \mathbb{R}^{N \times 1} \), which are further concatenated with the position \( \mathbf{P}\in \mathbb{R}^{N \times 3} \) to get final explicit physical attributes $\mathbf{F}^{\text{E}}$:
\begin{equation}
\mathbf{F}^{\text{E}} = \mathtt{Concat}(\mathbf{P}, \mathbf{R}, \mathbf{S}, \mathbf{O}) \in \mathbb{R}^{N \times 11}.
\end{equation}
Additionally, we assign each Gaussian with a learnable embedding \( \mathbf{F}^{\text{I}} \in \mathbb{R}^{N \times C} \), which serves as a query for aggregating image semantics and radar geometry. As a result, each Gaussian is represented as $\mathbf{G}=\{\mathbf{\mathbf{F}^{\text{E}},\mathbf{F}^{\text{I}}}\}$.

\begin{figure}[t]
    \centering
    \includegraphics[width=\linewidth]{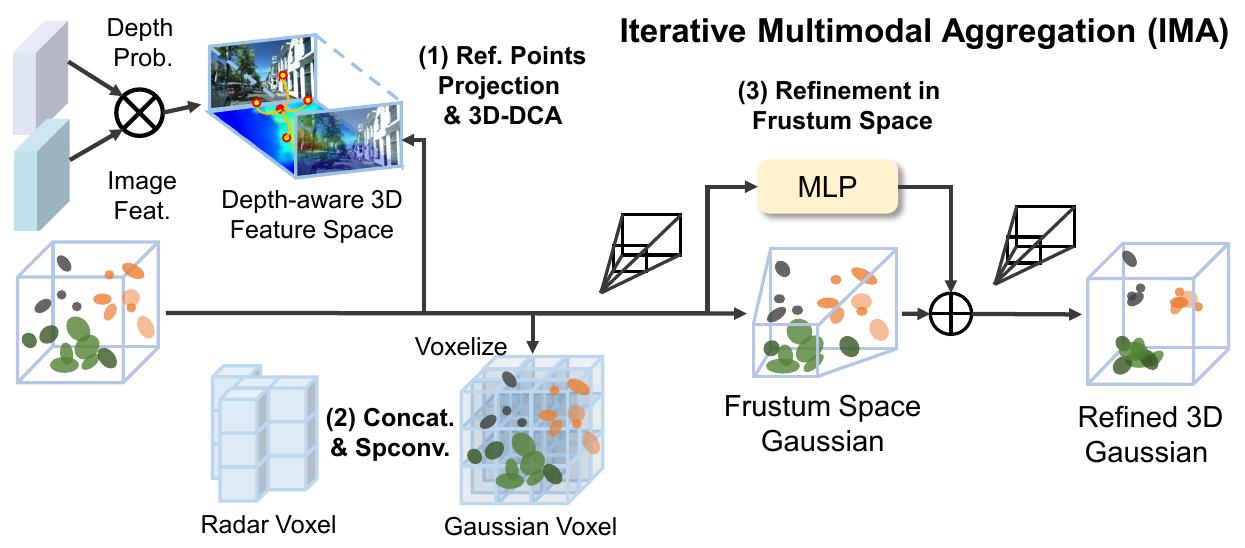}
    \caption{Procedure of Iterative Multimodal Aggregation (IMA). IMA involves the iterative aggregation of multi-modal features, followed by the updating of Gaussian locations within the frustum.}
    \label{fig:refinement}
\end{figure}

\begin{figure}[t]
    \centering
    \includegraphics[width=\linewidth]{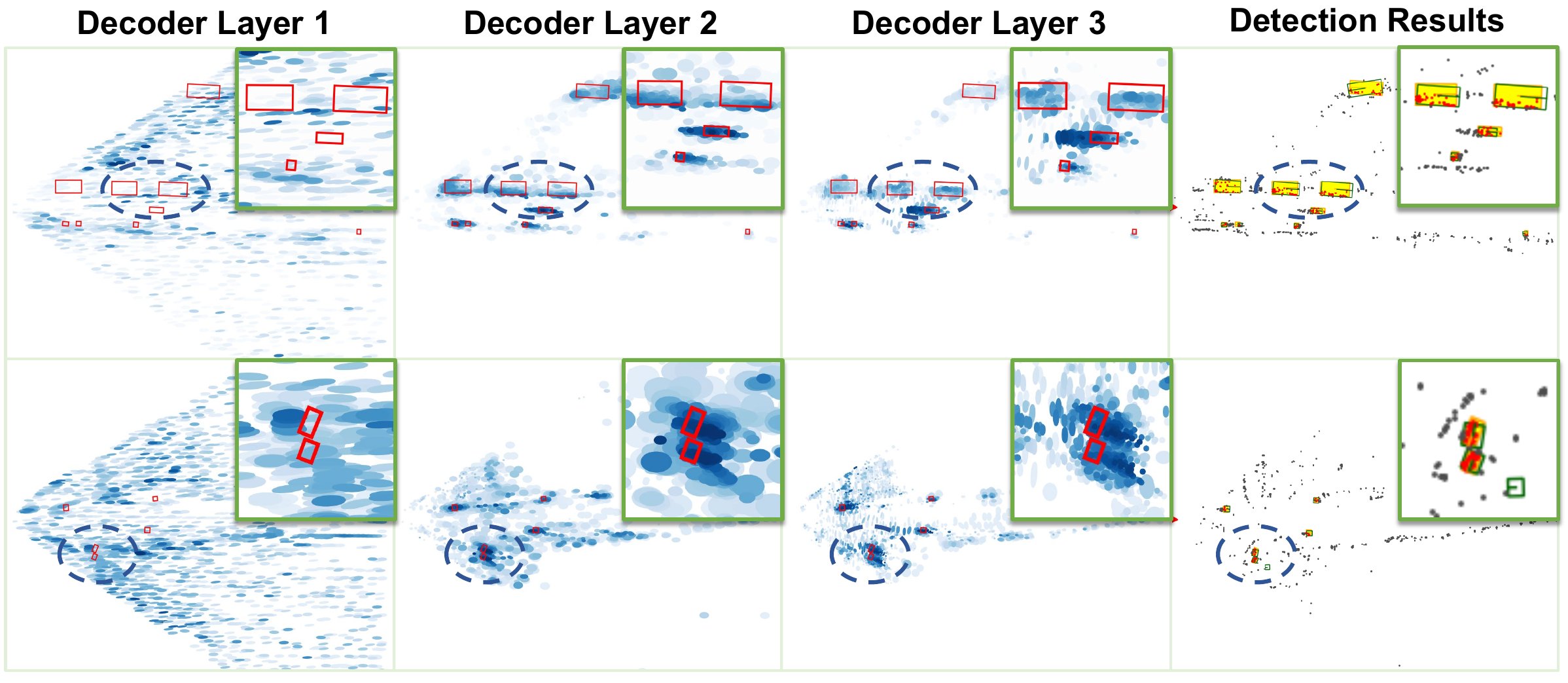}
    \caption{Dynamic Object Attention of RaGS. We visualize activated Gaussians (approximately 30\% of total) in the scene. RaGS focuses on sparse foreground objects while maintaining scene understanding.}
    \label{fig:gaussian}
\end{figure}

\noindent\textbf{Iterative Multimodal Aggregation (IMA).}
To enrich the Gaussians with multi-modal information and progressively guide them toward the foreground regions, we iteratively aggregate features from both image and radar domains. In particular, the spatial cues from radar serve as anchors for residual position updates, while the velocity features provide implicit guidance for gradual convergence. This iterative process outputs the final set of $M$ refined Gaussians, as illustrated in Fig.~\ref{fig:refinement}.

For semantic aggregation, instead of directly projecting Gaussians into 2D image space and applying 2D deformable attention for semantic aggregation, we first construct a depth-aware 3D image feature space out product image feature $\mathbf{{F}^{\text{2D}}}$ with depth probability $\mathbf{{D}^{\text{prob.}}}$. We then perform 3D deformable cross attention (3D-DCA) within it, enabling the Gaussians to interact with semantically and geometrically aligned image features in a spatially consistent manner. Specifically, given 3D Gaussian query $\mathbf{F}^{\text{I}}$ at location $\textbf{P}$, we aggregate image semantics by
\begin{equation}
\mathbf{F}^{\text{I}}  = \sum_{n=1}^{T} \mathbf{A}_n \mathbf{W} \cdot\phi(\mathbf{{F}^{\text{2D}}}\otimes\mathbf{{D}^{\text{prob.}}}, \mathcal{P}(\textbf{P}) + \Delta \textbf{q}),
\end{equation}
where $n$ indexes $T$ sampled offsets around $\textbf{P}$, $\mathcal{P}(\textbf{P})$ denotes projection function, $\Delta \textbf{q} \in \mathbb{R}^3$ denotes learnable offsets, and $\phi(\cdot)$ performs trilinear interpolation to extract features. The attention weights $\mathbf{A}_n \in [0,1]$ and projection matrix $\mathbf{W}$ are learned to guide the aggregation. The utilization of 3D deformable cross-attention (3D-DCA) \cite{SGDet3D} facilitates coherent interaction between Gaussians in 3D space and image features in the perspective view.

Consequently, we further enhance the implicit feature representation using sparse convolution, which integrates radar velocity as motion cues to refine the spatial distribution of Gaussians. Specifically, each optimized Gaussian point is treated as a point located at its center, and the resulting point cloud is voxelized into a sparse grid $\mathbf{V}^{\text{gs}} \in \mathbb{R}^{N \times C}$. Then, we incorporate radar pillars $\mathbf{F}^{\text{pillar}}$ consisting of rich geometry to enable synergistic multi-modal learning. Since radar pillars lack vertical resolution, we replicate each pillar along the height dimension to get $\mathbf{V}^{\text{radar}}\in \mathbb{R}^{(G\times Z) \times C}$, here $Z$ is the voxel number along the height dimension, which can be expressed as $\mathbf{V}^{\text{radar}}=\mathtt{Repeat}(\mathbf{F}^{\text{pillar}})$. These radar voxels are concatenated with $\mathbf{V}^{\text{gs}}$, and the RCS-velocity aware pillars provide clear physical meaning, thereby better assisting the Gaussians. By concatenating both modalities in a unified voxel space and applying sparse convolution, we treat each Gaussian as voxels and enable deep interaction between radar geometry and image semantics. This process is formulated as
\begin{equation}
\mathbf{F}^{\text{I}} \leftarrow \mathbf{V}^{\text{gs}} = \mathtt{Spconv}(\mathtt{Concat}(\mathbf{V}^{\text{gs}}, \mathbf{V}^{\text{radar}}))[:N],
\end{equation}
where $\mathtt{Spconv}$ denotes sparse convolution. Finally, to maintain consistency with the original Gaussian set, we only output former $N$ voxels using $[:N]$ and reassign to  $\mathbf{F}^{\text{I}}$, since the order remains unchanged. After updating the implicit feature, we begin explicit feature refinement as below.

Specifically, to achieve the finer foreground modeling required for 3D object detection, we refine each Gaussian localization within the frustum, which allows a more precise match with the monocular settings. To achieve this, we first reproject each Gaussian into frustum space and then using an MLP conditioned on its implicit feature $\mathbf{F}^{\text{I}}$ and transformed position $\mathcal{P}(\mathbf{P})$ to predict a residual spatial offset as
\begin{equation}
\Delta \mathbf{p}= (\Delta h, \Delta w, \Delta d)=\mathtt{MLP}(\mathtt{Concat}(\mathbf{F}^{\text{I}},\mathcal{P}(\mathbf{P}))),
\end{equation}
where $\Delta h, \Delta w, \Delta d$ corresponds to vertical, horizontal, and depth adjustments in perspective view and $\mathcal{P}$ denotes the projection from 3D space to the image frustum. The predicted residual is then added to frustum-space location of each Gaussian before being transformed back to 3D space, which ensures consistency with preceding modules. This process is formulated as
\begin{equation}
\mathbf{P} \leftarrow \mathcal{P}^{-1}(\mathcal{P}(\mathbf{P})+\Delta \mathbf{p}).
\end{equation}
This iterative process, depicted in Fig. \ref{fig:gaussian}, progressively refines Gaussian localization for sparse objects, in contrast to the fixed grid sampling of previous BEV-based models, which suffer from inflexibility and inefficiency due to the inevitable excessive background aggregation. All other Gaussian attributes are also updated simultaneously.

\begin{table}[t]
    \belowrulesep=0pt
    \aboverulesep=0pt
    \centering
    \footnotesize
    \setlength{\tabcolsep}{5pt}
    \renewcommand\arraystretch{1.05} %增加表格行距
    \begin{tabular}{c|c|c|cc}
        \toprule[1pt]
        % \makebox[2.5cm]{\centering Approach} & \makebox[1.2cm]{Res.} & \makebox[0.8cm]{\centering Input} & \makebox[1.0cm]{mAP(\%)} & \makebox[1.0cm]{ODS(\%)} \\
        % \midrule
        % PointPillars-128 (CVPR 2019) \cite{PointPillars} & L & 61.15 & 55.54 \\
        % PointPillars-32 (CVPR 2019) \cite{PointPillars} & L & 57.24 & 52.66 \\
        % \midrule
        % PointPillars (CVPR 2019) \cite{PointPillars} & - & R & 23.82 & 37.20 \\
        % RadarPillarNet (TIM 2023)  & - & R & 24.88 & 37.81 \\
        % \midrule
        % Lift, Splat, Shoot (ECCV 2020) \cite{LSS} & 544×960 & C & 22.44 & 26.01 \\
        % BEVFormer (ECCV 2022) \cite{BEVFormer}  & 544×960 & C & 26.49 & 28.10 \\
        % PanoOcc (CVPR2024) \cite{panoocc} & 544×960 & C & 29.17 & 28.55 \\
        % \midrule
        % BEVFusion (ICRA 2023) \cite{BEVFusion}  & 544×960 & C+R & 33.95 & 43.00 \\
        % RCFusion (TIM 2023) \cite{RCFusion}  & 544×960 & C+R & \underline{34.88} & \underline{41.53} \\
        % \rowcolor{gray!20} RaGS (\textbf{Ours}) & 544×960 & C+R & \textbf{35.88} & \textbf{43.45} \\
        \makebox[2.5cm]{\centering Approach} & \makebox[1.2cm]{Res.} & \makebox[0.8cm]{\centering Input} & \makebox[1.0cm]{mAP(\%)} & \makebox[1.0cm]{ODS(\%)} \\
        % \midrule
        % PointPillars-128 & L & 61.15 & 55.54 \\
        % PointPillars-32 & L & 57.24 & 52.66 \\
        \midrule
        PointPillars & - & R & 23.82 & 37.20 \\
        RadarPillarNet & - & R & 24.88 & 37.81 \\
        \midrule
        Lift, Splat, Shoot & 544×960 & C & 22.44 & 26.01 \\
        BEVFormer & 544×960 & C & 26.49 & 28.10 \\
        PanoOcc & 544×960 & C & 29.17 & 28.55 \\
        \midrule
        BEVFusion & 544×960 & C+R & \cellcolor{third}33.95 & \cellcolor{second}\underline{43.00} \\
        RCFusion  & 544×960 & C+R & \cellcolor{second}\underline{34.88} & \cellcolor{third}41.53 \\
        \rowcolor{gray!20} RaGS (\textbf{Ours}) & 544×960 & C+R & \cellcolor{best}\textbf{35.88} & \cellcolor{best}\textbf{43.45} \\
        \bottomrule[1pt]
    \end{tabular}
    \caption{Comparison on the OmniHD-Scenes \cite{OmniHD} \emph{test} set.}
    \label{tab:OmniHD-Scenes}
\end{table}

\begin{table}[t]
    \belowrulesep=0pt
    \aboverulesep=0pt
    \centering
    \footnotesize
    \setlength{\tabcolsep}{2pt}
    \renewcommand\arraystretch{1.05} %增加表格行距
    \begin{tabular}{c|c|c|c}
        \toprule[1.0pt]
        \makebox[4.3cm]{\centering Approach} & \makebox[1cm]{\centering Input} & \makebox[1.20cm]{$\text{AP}_\text{3D}$ (\%)} & \makebox[1.20cm]{$\text{AP}_\text{BEV}$ (\%)}
        \\ \midrule
        ImVoxelNet (WACV 2022) \cite{imvoxelnet} & C & 14.96 & 17.12 \\
        \midrule
        RadarPillarNet (TIM 2023) \cite{RCFusion} & R & 30.37 & 39.24 \\
        SMURF (TIV 2023) \cite{SMURF} & R & 32.99 & 40.98 \\
        \midrule
        FUTR3D (CVPR 2023) \cite{FUTR3D}  & R+C & 32.42 & 37.51 \\
        BEVFusion (ICRA 2023) \cite{BEVFusion}  & R+C & 32.71 & 41.12 \\
        LXL (TIV 2024)  \cite{LXL} & R+C & 36.32 & 41.20 \\ 
        RCFusion (TIM 2023) \cite{RCFusion}  & R+C & 33.85 & 39.76 \\
        LXLv2 (RAL 2025) \cite{LXLv2} & R+C & 37.32 & 42.35 \\ 
        UniBEVFusion (ICRA 2025) \cite{UniBEVFusion}  & R+C & 37.76 & 42.92 \\
        MSSF (TITS 2025) \cite{MSSF}  & R+C & \cellcolor{third}37.97 & 43.11 \\
        HGSFusion (AAAI 2025) \cite{HGSFusion} & R+C & 37.21 & \cellcolor{third}43.23 \\
        SGDet3D (RAL 2025) \cite{SGDet3D}  & R+C & \cellcolor{second}\underline{41.82} & \cellcolor{second}\underline{47.16} \\
        \rowcolor{gray!20} RaGS (\textbf{Ours}) & R+C & \cellcolor{best}\textbf{41.95} & \cellcolor{best}\textbf{51.04} \\ 
        \bottomrule[1.0pt]
    \end{tabular}
    \caption{Comparison on the \emph{test} set of TJ4DRadSet \cite{TJ4D}.}
    \label{tab:TJ4D}
\end{table}

\begin{table}[t]
    \belowrulesep=0pt
    \aboverulesep=0pt
    \centering
    \footnotesize
    \renewcommand\arraystretch{1.2}
    \setlength{\tabcolsep}{2.0pt}
    \begin{tabular}{c|cc|cc}
        \toprule[1.0pt]
        \multirow{2}{*}{\makebox[3.5cm]{Models}} & 
        \multicolumn{2}{c|}{\makebox[2.0cm]{View-of-Delft\cite{VoD}}} & 
        \multicolumn{2}{c}{\makebox[2.0cm]{TJ4DRadSet\cite{TJ4D}}} \\ 
        \cmidrule(lr){2-3} \cmidrule(lr){4-5}
        & \makebox[1.0cm]{$\text{mAP}_\text{EAA}$} & \makebox[1.0cm]{$\text{mAP}_\text{DC}$} 
        & \makebox[1.0cm]{$\text{mAP}_\text{3D}$} & \makebox[1.0cm]{$\text{mAP}_\text{BEV}$} \\ 
        \midrule
        FUTR3D (CVPR2023) \cite{FUTR3D} & \cellcolor{third}49.03 & \cellcolor{third}69.32 & \cellcolor{second}32.42 & \cellcolor{second}37.51 \\
        RaCFormer (CVPR2025) \cite{racformer} & \cellcolor{second}54.44 & \cellcolor{second}78.57 & - & - \\
        \rowcolor{gray!20} RaGS (\textbf{Ours}) & \cellcolor{best}\textbf{61.86} & \cellcolor{best}\textbf{81.63} & \cellcolor{best}\textbf{41.95} & \cellcolor{best}\textbf{51.04} \\
        \bottomrule[1.0pt]
    \end{tabular}
    \caption{Comparison with DETR-based models.}
    \label{tab:DETR}
\end{table}

\begin{figure*}[t]
    \centering
    \includegraphics[width=0.97\linewidth]{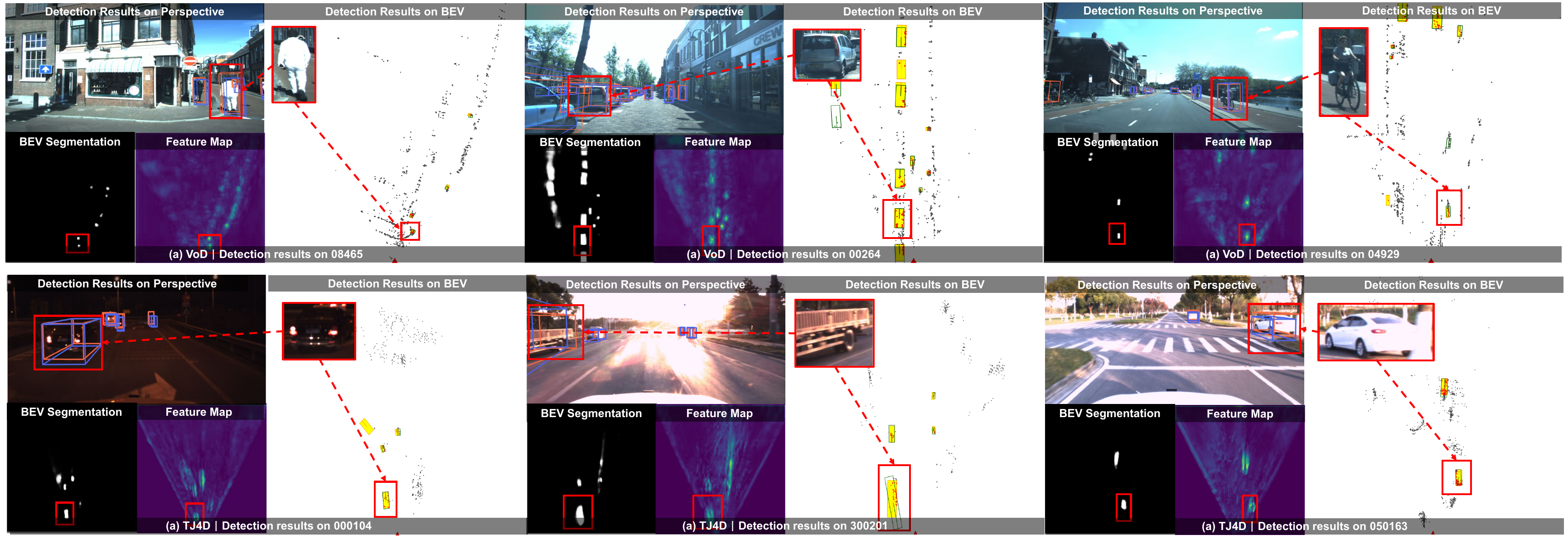}
    \caption{Visualization results on the VoD validation set (first row) and TJ4DRadSet test set (second row) . Each figure corresponds to a frame. Orange and yellow boxes represent ground-truths in the perspective and bird’seye views, respectively. Green and blue boxes indicate predicted results. Zoom in for better view.}
    \label{fig:visualization}
\end{figure*}

\begin{table*}[ht]
    \belowrulesep=0pt
    \aboverulesep=0pt
    \centering
    \footnotesize
    \setlength{\tabcolsep}{4.5pt}
    \renewcommand\arraystretch{1.05} %增加表格行距
	% \resizebox{1.0\linewidth}{!}{
    \begin{tabular}{l|c|cccc|cccc|c}
        \toprule[1.0pt]
        \makebox[5.00cm]{\multirow{2}{*}[-0.7ex]{\centering Approach}} & \makebox[1.00cm]{\multirow{2}{*}[-0.7ex]{\centering Input}} & \multicolumn{4}{c|}{Entire Annotated Area (\%)} & \multicolumn{4}{c|}{Driving Corridor (\%)} & \multirow{2}{*}[-0.7ex]{\centering FPS} \\ 
        \cline{3-6} \cline{7-10}
         &  & \makebox[0.90cm]{Car} & \makebox[0.90cm]{Ped} & \makebox[0.90cm]{Cyclist} & \makebox[0.90cm]{mAP} & \makebox[0.90cm]{Car} & \makebox[0.90cm]{Ped} & \makebox[0.90cm]{Cyclist} & \makebox[0.90cm]{mAP} \\ \midrule
        ImVoxelNet (WACV 2022) \cite{imvoxelnet} & C & 19.35 & 5.62 & 17.53 & 14.17 & 49.52 & 9.68 & 28.97 & 29.39 & 11.1\\
        \midrule
        PillarNeXt (CVPR 2023) \cite{pillarnext} & R & 30.81 & 33.11 & 62.78 & 42.23 & 66.72 & 39.03 & 85.08 & 63.61 & -\\
        % PointPillars (CVPR 2019) \cite{PointPillars} & R & 37.06 & 35.04 & 63.44 & 45.18 & 70.15 & 47.22 & 85.07 & 67.48 & 113.9 \\
        RadarPillarNet (TIM 2023) \cite{RCFusion} & R & 39.30 & 35.10 & 63.63 & 46.01 & 71.65 & 42.80 & 83.14 & 65.86 & 98.8 \\
        CenterPoint (CVPR 2021) \cite{centerpoint} & R & 35.84 & 41.03 & 67.11 & 47.99 & 70.65 & 50.14 & 85.67 & 68.82 & 38.3\\
        VoxelNeXt (CVPR 2023) \cite{VoxelNeXt} & R & 36.98 & 42.37 & 68.15 & 49.17 & 70.95 & 51.85 & 87.33 & 70.04 & 31.6\\
        % LXL - R (TIV 2023) \cite{LXL} & R & 32.75 & 39.65 & 68.13 & 46.84 & 70.26 & 47.34 & 87.93 & 68.51 & 44.7\\
        SMURF (TIV 2023) \cite{SMURF} & R & 42.31 & 39.09 & 71.50 & 50.97 & 71.74 & 50.54 & 86.87 & 69.72 & -\\
        SCKD (AAAI 2025) \cite{SCKD} & R & 41.89 & 43.51 & 70.83 & 52.08 & 77.54 & 51.06 & 86.89 & 71.80 & 39.3\\
        % \midrule
        % InterFusion (IROS 2023) \cite{InterFusion} & R+L & 66.50 & 64.50 & 78.50 & 69.83 & 90.70 & 72.00 & 88.70 & 83.80 & - \\  
        % L4DR (AAAI 2025) \cite{L4DR} & R+L & 69.10 & 66.20 & 82.80 & 72.70 & 90.80 & 76.10 & 95.50 & 87.47 & - \\  
        \midrule
        FUTR3D (CVPR 2023) \cite{FUTR3D} & R+C & 46.01 & 35.11 & 65.98 & 49.03 & 76.98 & 43.10 & 86.19 & 69.32 & 7.3 \\
        BEVFusion (ICRA 2023) \cite{BEVFusion} & R+C & 37.85 & 40.96 & 68.95 & 49.25 & 70.21 & 45.86 & \cellcolor{third}89.48 & 68.52 & 7.1\\
        RCFusion (TIM 2023) \cite{RCFusion} & R+C & 41.70 & 38.95 & 68.31 & 49.65 & 71.87 & 47.50 & 88.33 & 69.23 & \cellcolor{third}9.0\\
        % TL-4DRCF (Sensors 2024) \cite{TL-4DRCF} & R+C & 43.71 & 40.11 & 64.22 & 49.35 & 79.49 & 53.76 & 76.50 & 69.92 & -\\
        RCBEVDet (CVPR 2024) \cite{RCBEVDet} & R+C & 40.63 & 38.86 & 70.48 & 49.99 & 72.48 & 49.89 & 87.01 & 69.80 & -\\
        ZFusion (CVPR 2025) \cite{zfusion} & R+C  & 43.89  & 39.48  & 70.46  & 51.28  & 79.51  & 52.95 & 86.37 & 72.94 & - \\
        % UniBEVFusion (ICRA 2025) \cite{UniBEVFusion} & R+C & 42.22 & 47.11 & 72.94 & 54.09 & 72.10 & 57.71 & \cellcolor{best}\textbf{93.29} & 74.37 & -\\
        % RaCFormer (CVPR 2025) \cite{racformer} & R+C & 47.30 & 46.21 & 69.80 & 54.44 & \cellcolor{best}\textbf{89.26} & 56.78 & \cellcolor{third}89.67 & \cellcolor{second}\underline{78.57} & - \\
        IS-Fusion (CVPR 2024) \cite{IS-Fusion} & R+C & \cellcolor{third}48.57 & 46.17 & 68.48 & \cellcolor{third}54.40 & \cellcolor{third}80.42 & 55.50 & 88.33 & \cellcolor{third}74.75 & -\\
        LXL (TIV 2024)  \cite{LXL} & R+C & 42.33 & \cellcolor{third}49.48 & \cellcolor{best}\textbf{77.12} & 56.31 & 72.18 & \cellcolor{third}58.30 & 88.31 & 72.93 & 6.1\\ 
        % LXLv2 (RAL 2025) \cite{LXLv2} & R+C & 47.81 & 49.30 & \textbf{77.15} & 58.09 & - & - & - & - & 6.5\\ 
        % HGSFusion (AAAI 2025) \cite{HGSFusion} & R+C & 51.67 & \textbf{52.64} & 72.58 & 58.96 & \textbf{88.28} & \textbf{62.61} & 87.49 & \underline{79.46} & 3.2 \\
        SGDet3D (RAL 2025) \cite{SGDet3D} & R+C & \cellcolor{second}\underline{53.16} & \cellcolor{second}\underline{49.98} & \cellcolor{third}76.11 & \cellcolor{second}\underline{59.75} & \cellcolor{second}\underline{81.13} & \cellcolor{second}\underline{60.91} & \cellcolor{second}\underline{90.22} & \cellcolor{second}\underline{77.42} & \cellcolor{second}\underline{9.2}\\
        % MSSF-V (TITS 2025) & R+C & 52.53 & 51.58 & 75.77 & 59.96 & 89.08 & 66.78 & 88.10 & 81.32 & 10.3\\
        % MSSF-PP (TITS 2025) & R+C & 60.96 & 51.28 & 77.69 & 63.31 & 90.60 & 60.39 & 88.35 & 79.78 & 13.9\\
        % HyDRa (ICRA 2025) & R+C & 52.83 & \textbf{56.57} & 73.25 & \underline{60.88} & 80.65 & \textbf{62.90} & 87.43 & 76.99 & - \\
        % \rowcolor{gray!20} PointPillars (CVPR 2019) \cite{PointPillars} & L & 60.48 & 45.78 & 69.16 & 58.47 & 90.27 & 61.81 & 87.13 & 79.74 & 63.8 \\
        % \rowcolor{gray!20} RaGS (\textbf{Ours}) & R+C & \textbf{57.71} & 49.06 & 75.86 & \textbf{60.89} & \textbf{83.18} & \underline{60.12} & 87.83 & \underline{77.04} & 10.5\\ 
        \rowcolor{gray!20} RaGS (\textbf{Ours}) & R+C & \cellcolor{best}\textbf{58.15} & \cellcolor{best}\textbf{50.81} & \cellcolor{second}\underline{76.62} & \cellcolor{best}\textbf{61.86} & \cellcolor{best}\textbf{88.15} & \cellcolor{best}\textbf{61.67} & \cellcolor{best}\textbf{95.07} & \cellcolor{best}\textbf{81.63} & \cellcolor{best}\textbf{10.5}\\ 
        \bottomrule[1.0pt]
    \end{tabular}
    \caption{Comparison with state-of-the-art approaches on the \emph{val} set of View-of-Delft \cite{VoD}.}
    \label{tab:VoD}
\end{table*}

\noindent\textbf{Multi-level Gaussian Fusion (MGF).}
After obtaining multiple Gaussian representations that capture different levels of semantic and geometric abstraction, we further integrate them to form a unified, multi-level feature representation for subsequent rendering and detection. Through IMA, we obtain Gaussians set consisting of $M$ refined Gaussians $\mathcal{G} = \{ \mathbf{G}_i \}_{i=1}^M$. 
Each $\mathbf{G}_i$ has $N$ Gaussian, inheriting its explicit attributes $(\mathbf{P}_i, \mathbf{R}_i, \mathbf{S}_i, \mathbf{O}_i)$ and implicit feature $\mathbf{F}^{\text{I}}_i$ from the previous stage, which can be formally represented as $\mathbf{G}_i = (\boldsymbol{\mu}_i, \boldsymbol{\Sigma}_i, \alpha_i, \mathbf{f}_i)$, 
where the Gaussian center $\boldsymbol{\mu}_i$ corresponds to $\mathbf{P}_i$, the covariance $\boldsymbol{\Sigma}_i$ is derived from the rotation $\mathbf{R}_i$ and scale $\mathbf{S}_i$, $\alpha_i$ denotes opacity (from $\mathbf{O}_i$), and $\mathbf{f}_i$ is the learned feature embedding.

For each of $N$ Gaussian within $\mathbf{G}_i$, it contributes to BEV cells through differentiable Gaussian Splatting. Specifically, given a BEV pixel at position $\mathbf{q}=(x,y)$, its feature value is computed as the weighted accumulation of all projected Gaussians, formulated as
\begin{equation}
\sum_{n=1}^{N} \alpha_n 
\exp\!\left(-\tfrac{1}{2}(\mathbf{q}-\boldsymbol{\mu}_{n,xy})^{\!\top}\!\boldsymbol{\Sigma}_{n,xy}^{-1}(\mathbf{q}-\boldsymbol{\mu}_{n,xy})\right)\mathbf{f}_n,
\label{eq:gsplat}
\end{equation}
where $\boldsymbol{\mu}_{n,xy}$ and $\boldsymbol{\Sigma}_{n,xy}$ are the mean and covariance of the Gaussian projected onto the BEV plane. 
This operation defines a smooth, differentiable mapping from the 3D Gaussian field to the 2D BEV domain, implemented efficiently via a CUDA-based rasterizer \cite{Gaussianformer}. The implicit features of the last $L~(L\!\leq\!M)$ Gaussians are rasterized into multi-level BEV-aligned feature maps $\{\mathbf{F}^{(l)} \in \mathbb{R}^{X \times Y \times C}\}$ as
\begin{equation}
\mathbf{F}^{(l)} = \mathtt{Rasterize}(\mathbf{G}_{M-L+l}), \quad l = 1, \ldots, L,
\end{equation}
where $(X,Y)$ denotes the BEV resolution, $\mathtt{Rasterize}$ denotes the rasterizer. These rendered maps form a hierarchical Gaussian representation across spatial scales. We then fuse the multi-level features via convolution as
\begin{equation}
\mathbf{F}^{\text{gs}} = \mathtt{Conv}\!\left(\mathtt{Concat}(\mathbf{F}^{(1)}, \ldots, \mathbf{F}^{(L)})\right).
\end{equation}
The fused Gaussian feature $\mathbf{F}^{\text{gs}}$ is further enhanced by radar-derived sparse pillars $\mathbf{F}^{\text{pillar}}$ through a cross-modal fuser, formulated as $\mathbf{F}^{\text{BEV}} = \mathtt{CMF}(\mathbf{F}^{\text{gs}}, \mathbf{F}^{\text{pillar}})$, where $\mathtt{CMF}$ denotes a lightweight module composed of concatenation and convolution layers. As a result, our MGF generates a hierarchical representation that compensates for the sparse and low-resolution nature of the Gaussian from IMA.

Additionally, we render final-layer Gaussians into depth supervised by LiDAR, and predict a BEV segmentation map from fused BEV features with occupancy guidance. Crucially, while MGF rasterizes the Gaussian field into BEV space, the Gaussians themselves are dynamically refined toward object-centric regions, which allows adaptive aggregation that prioritizes sparse foreground objects and mitigates the background redundancy of grid-based BEV methods.
% After iterative cross-modal aggregation through IMA, we obtain a set of $M$ refined Gaussians, which are then rasterized into multi-level BEV-aligned feature maps via Gaussian splatting at different aggregation stages. This generates a hierarchical representation that compensates for the sparse and low-resolution nature of the initial Gaussian anchors. The resulting BEV features from $L(L<=M)$ levels are fused to enable robust cross-modal integration and are subsequently forwarded to the detection head for final 3D object prediction. During this process, we render the final-layer Gaussians into a perspective depth map, optimize opacity using LiDAR depth, and use the fused BEV features to predict a segmentation map in the bird's-eye view, supervised by the occupancy map.

\subsection{Detection Head and Loss Function}
On one hand, following \cite{SGDet3D}, we apply the depth loss $\mathcal{L}_{\text{depth}}$ and the perspective segmentation loss $\mathcal{L}_{\text{seg}}$ to supervise the raw feature extraction for pretraining, expressed as 
\begin{equation}
\mathcal{L}_{\text{pretrain}} = \mathcal{L}_{\text{depth}} + \mathcal{L}_{\text{seg}}.
\end{equation}
On the other hand, we adopt the 3D object detection loss $\mathcal{L}_{\text{det}}$ following \cite{RCFusion}, along with auxiliary rendering losses: the rendered depth loss in the perspective view $\mathcal{L}_{\text{depth\_render}}$, and the semantic segmentation loss in the BEV $\mathcal{L}_{\text{seg\_render}}$. The total joint training loss is defined as
\begin{equation}
\mathcal{L}_{\text{total}} = \mathcal{L}_{\text{det}} + \lambda(\mathcal{L}_{\text{depth\_render}} + \mathcal{L}_{\text{seg\_render}}),
\end{equation}
where $\lambda$ is the hyperparameter balancing detection and auxiliary tasks. In this work, we simply set $\lambda=0.1$. The perspective and BEV segmentation ground-truths can be simply generated through detectron2 \cite{wu2019detectron2} and 3D bounding boxes. See supplementary material for more details. 
% The ground-truths of perspective segmentation are simply generated through detectron2 \cite{wu2019detectron2}, while BEV segmentation can easily inferred from 3d bounding boxes. See supplementary material for losses ablation study and more details. 
% As 3D object detection task is relatively sparse than reconstruction tasks, auxiliary rendering losses are essential to guide Gaussians in learning accurate opacity at object regions, enhancing their saliency in feature maps. See Table \ref{tab:ablation_aux} for detailed ablation.

\subsection{Implementation Details}
\textbf{Datesets and Evaluation Metrics.} 
We evaluate our model on three 4D radar benchmarks: VoD \cite{VoD}, TJ4DRadSet \cite{TJ4D}, and OmniHD-Scenes \cite{OmniHD}. The VoD dataset, collected in Delft, includes 5,139 training frames, 1,296 validation frames, and 2,247 test frames. The TJ4DRadSet, captured in Suzhou, contains 7,757 frames with 5,717 for training and 2,040 for testing, posing challenges due to complex scenarios such as nighttime and glare. The OmniHD-Scenes dataset features multimodal data from six cameras, six 4D radars, and a 128-beam LiDAR, with 11,921 keyframes annotated, consisting of 8,321 for training, 3,600 for testing.

We follow the official evaluation protocols for each dataset. For VoD, we report 3D Average Precision (AP) across the full area and drivable corridor. For TJ4DRadSet, we evaluate 3D AP and BEV AP within 70 meters of the radar origin. For OmniHD-Scenes, we use mean Average Precision (mAP) and the OmniHD-Scenes Detection Score (ODS) for detection within a ±60 m longitudinal and ±40 m lateral range around the ego vehicle.

\noindent \textbf{Network Settings and Training details.}
For all datasets, the final voxel is set to a cube of size 0.32 m, and the image sizes are 800×1280 for VoD, 640×800 for TJ4DRadSet, and 544×960 for OmniHD-Scenes. Anchor size and point cloud range are kept as in \cite{OmniHD}. The models are trained on 4 NVIDIA GeForce RTX 4090 GPUs with a batch size of 4 per GPU. AdamW is used as the optimizer, with 12 epochs for pretraining and 24 epochs for joint training.
% Unless otherwise specified, other training settings follow those of \cite{SGDet3D}.

\begin{table}[t]
    \belowrulesep=0pt
    \aboverulesep=0pt
    \centering
    \footnotesize
    \setlength{\tabcolsep}{11.5pt}
    \renewcommand\arraystretch{1.05}
    \begin{tabular}{c|c|c|c|c|c}
        \toprule[1.0pt] 
        \makebox[0.5cm]{Baseline} & \makebox[0.5cm]{w/ FLI} & \makebox[0.5cm]{w/ IMA} & \makebox[0.5cm]{w/ MGF} & \makebox[0.5cm]{$\text{mAP}_\text{EAA}$} & \makebox[0.5cm]{$\text{mAP}_\text{DC}$} \\
        \midrule
        % \checkmark &  &  &  & 10.76 & 20.33 \\
        \checkmark &  &  &  & 55.33 & 72.32 \\
        \checkmark & \checkmark &  &  & 57.40 & 75.80 \\
        \checkmark & \checkmark & \checkmark &  & 59.12 & 76.68 \\
        \checkmark & \checkmark & \checkmark & \checkmark & \textbf{59.45} & \textbf{76.98} \\
        \bottomrule[1.0pt]
    \end{tabular}
    \caption{Overall ablation of RaGS.}
    \label{tab:ablation_ALL}
    
\end{table}
\begin{table}[t]
    \belowrulesep=0pt
    \aboverulesep=0pt
    \centering
    \footnotesize
    \setlength{\tabcolsep}{7.5pt}
    \renewcommand\arraystretch{1.05}
    \begin{tabular}{c|c|c|c|c|c}
        \toprule[1.0pt]
        \makebox[1.1cm]{\multirow{2}{*}{Baseline}} & \multicolumn{3}{c|}{FLI} & \makebox[0.8cm]{\multirow{2}{*}{$\text{mAP}_\text{EAA}$}} & \makebox[0.8cm]{\multirow{2}{*}{$\text{mAP}_\text{DC}$}} \\
        \cmidrule(lr){2-4}
        & \makebox[0.8cm]{Frustum} & \makebox[0.8cm]{Radar} & \makebox[0.8cm]{Depth} & & \\
        \midrule
        \checkmark &  &  &  & 55.33 & 72.32 \\
        \checkmark & \checkmark &  &  & 56.26 & 73.10 \\
        \checkmark & \checkmark & \checkmark &  & 56.72 & 75.78 \\
        \checkmark & \checkmark & \checkmark & \checkmark & \textbf{57.40} & \textbf{75.80} \\
        \bottomrule[1.0pt]
    \end{tabular}
    \caption{Ablation study of FLI on the VoD dataset.}
    \label{tab:ablation_FLI}
\end{table}

\section{Experiments}
\subsection{3D Object Detection Results}
\noindent\textbf{Results on OmniHD-Scenes Dataset.}
We evaluated the performance of RaGS in a surround-view scenario, as shown in Table \ref{tab:OmniHD-Scenes}. RaGS consistently outperforms traditional fixed-grid models RCFusion and BEVFusion on the mAP and ODS. This underscores the advantage of utilizing Gaussian representations for the fusion of 4D radar and surround-view data in the context of 3D object detection. For multi-view setting, our ongoing work is exploring more elegant initialization prior like VGGT \cite{VGGT}.

\begin{table}[t]
    \belowrulesep=0pt
    \aboverulesep=0pt
    \centering
    \footnotesize
    \setlength{\tabcolsep}{7.5pt}
    \renewcommand\arraystretch{1.05}
    \begin{tabular}{c|c|c|c|c|c}
        \toprule[1.0pt]
        \makebox[1.1cm]{\multirow{2}{*}{FLI}} & \multicolumn{3}{c|}{IMA} & \makebox[0.8cm]{\multirow{2}{*}{$\text{mAP}_\text{EAA}$}} & \makebox[0.8cm]{\multirow{2}{*}{$\text{mAP}_\text{DC}$}} \\
        \cmidrule(lr){2-4}
        & \makebox[0.8cm]{3D-DCA} & \makebox[0.8cm]{Pillars} & \makebox[0.8cm]{Frustum} & & \\
        \midrule
        \checkmark &  &  &  & 57.40 & 75.80 \\
        \checkmark & \checkmark &  &  & 58.79 & 76.16 \\
        \checkmark & \checkmark & \checkmark &  & 58.93 & 76.35 \\
        \checkmark & \checkmark & \checkmark & \checkmark & \textbf{59.45} & \textbf{76.98} \\
        \bottomrule[1.0pt]
    \end{tabular}
    \caption{Ablation study of IMA on the VoD dataset.}
    \label{tab:ablation_IMA}
\end{table}

\begin{table}[t]
    \belowrulesep=0pt
    \aboverulesep=0pt
    \centering
    \footnotesize
    \renewcommand\arraystretch{1.05}
    \begin{tabular}{c|ccc|c|c|c}
        \toprule[1.0pt]
        % \makebox[0.8cm]{\multirow{2}{*}{$L$}} & \multicolumn{3}{c|}{MGF Fusion Level} & \makebox[0.8cm]{\multirow{2}{*}{$\text{mAP}_\text{EAA}$}} & \makebox[0.8cm]{\multirow{2}{*}{$\text{mAP}_\text{DC}$}} & \makebox[0.8cm]{\multirow{2}{*}{GFLOPS}} \\
        % \cmidrule(lr){2-4}
        % & \makebox[0.5cm]{Car} & \makebox[0.5cm]{Ped} & \makebox[0.5cm]{Cyc} & & \\
        % \midrule
        % None & 52.03 & 48.84 & 76.48 & 59.12 & 75.68 & 579.87\\
        % 2 & \textbf{52.94} & 48.31 & \textbf{77.10} & \textbf{59.45} & \textbf{76.98} & 639.82\\
        % 3 & 52.29 & \textbf{48.46} & 76.79 & 59.18 & 76.63 & 689.48\\
        % \midrule
        
        % \makebox[0.8cm]{\multirow{2}{*}{$M$}} & \multicolumn{3}{c|}{IMA Aggr. Number} & \makebox[0.8cm]{\multirow{2}{*}{$\text{mAP}_\text{EAA}$}} & \makebox[0.8cm]{\multirow{2}{*}{$\text{mAP}_\text{DC}$}} & \makebox[0.8cm]{\multirow{2}{*}{FLOPS}}\\
        % \cmidrule(lr){2-4}
        % & \makebox[0.5cm]{Car} & \makebox[0.5cm]{Ped} & \makebox[0.5cm]{Cyc} & & \\
        % \midrule
        % 1 & 49.70 & 48.06 & 75.92 & 57.89 & 75.01 & 538.36 \\
        % 2 & 50.61 & 48.23 & 76.42 & 58.42 & \textbf{77.45} & 572.03 \\
        % 3 & 52.94 & \textbf{48.31} & \textbf{77.10} & 59.45 & 76.98 & 605.70\\
        % 4 & \textbf{53.62} & 48.14 & 76.90 & \textbf{59.55} & 76.88 & 639.82\\
        % \midrule
        
        \makebox[0.8cm]{\multirow{2}{*}{$N$}} & \multicolumn{3}{c|}{Gaussian Anchor} & \makebox[0.8cm]{\multirow{2}{*}{$\text{mAP}_\text{EAA}$}} & \makebox[0.8cm]{\multirow{2}{*}{$\text{mAP}_\text{DC}$}} & \makebox[0.8cm]{\multirow{2}{*}{FLOPS}} \\
        \cmidrule(lr){2-4}
        & \makebox[0.5cm]{Car} & \makebox[0.5cm]{Ped} & \makebox[0.5cm]{Cyc} & & \\
        \midrule
        3200 & 52.24 &  41.79 & 70.29 & 54.77 & 76.13 & 599.39 \\
        6400 & 52.23 &  41.92 & 76.35 & 56.83 & 76.22 & 611.95\\
        12800 & \textbf{52.94} & 48.31 & \textbf{77.10} & 59.45 & \textbf{76.98} & 639.82\\
        19200 & 52.63 & \textbf{48.84} & 76.94 & \textbf{59.47} & 76.70 & 666.06\\
        \bottomrule[1.0pt]
    \end{tabular}
    % \caption{Ablation of fusion level number $L$ in MGF, iterative aggregation number $M$ in IMA and anchor number $N$.}
    \caption{Ablation of anchor number $N$.}
    \label{tab:ablation_num}
\end{table}

\begin{figure}[t]
    \centering
    \includegraphics[width=\linewidth]{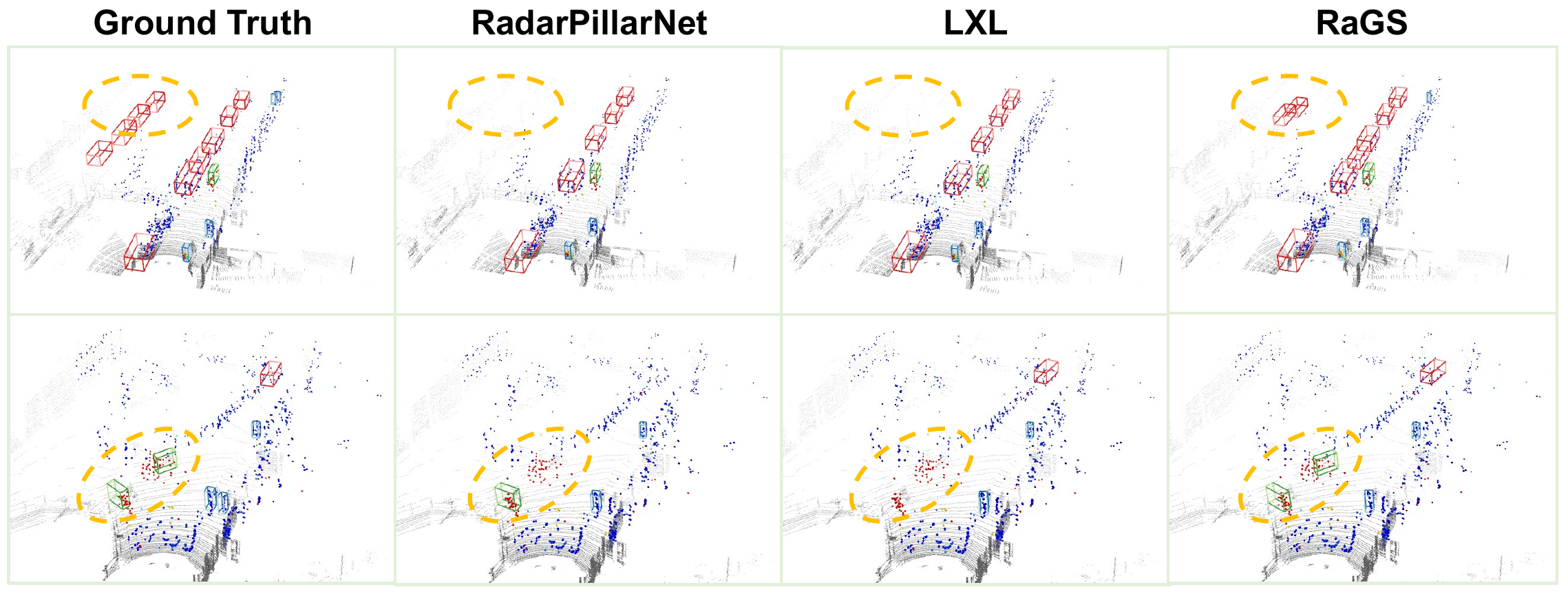}
    \caption{Qualitative comparison with state-of-the-art.}
    \label{fig:3Dcomparison}
\end{figure}

\noindent\textbf{Results on TJ4DRadSet and View-of-Delft Dataset.}
Compared to the VoD dataset, the TJ4DRadSet dataset \cite{TJ4D} presents greater challenges, including complex scenarios such as nighttime, under-bridge conditions, and the addition of a truck category with highly variable object sizes. Despite these difficulties, RaGS achieves state-of-the-art performance, with the highest $\text{mAP}_{\text{BEV}}$ of 51.04\% and the best $\text{mAP}_{\text{3D}}$ of 41.95\%, as shown in Table \ref{tab:TJ4D}. On the other side, Table \ref{tab:VoD} presents the 3D object detection results on the VoD validation set \cite{VoD}, where our RaGS consistently outperforms state-of-the-art 4D radar and camera fusion models. For cars, we achieve exceptional accuracy, leveraging the flexible occupancy capability of the Gaussian model and an effective fusion design. For pedestrians and cyclists, we also deliver near best performance. When compared to the strong BEV-based baseline LXL \cite{LXL}, our method shows significant improvements, with gains of 5.55\% in $\text{mAP}_{\text{EAA}}$ and 8.70\% in $\text{mAP}_{\text{DC}}$. Furthermore, it achieves a frame rate of 10.5 FPS, demonstrating its suitability for real-world applications. Visualization results of VoD and TJ4DRadSet are shown in the first and second row of Fig. \ref{fig:visualization}, respectively. Qualitative comparison are presented in Fig. \ref{fig:3Dcomparison}. Compared to DETR-based methods such as FUTR3D (49.03\%) and RaCFormer (54.44\%), as shown in Table \ref{tab:DETR}, our RaGS (61.86\%) demonstrates a significant performance advantage. Unlike implicit end-to-end query-based frameworks, the 3D Gaussian Splatting in RaGS offers explicit physical interpretability, serving as a multi-modal aggregator that fuses semantic and geometric features, leveraging radar for 3D perception and monocular cues for enriched observation.

\subsection{Ablation Study}
\noindent\textbf{Overall Ablation.}
Table \ref{tab:ablation_ALL} reports the overall ablation on RaGS, conducted with half training epochs for efficiency. Starting from the baseline \cite{Gaussianformer} with 4D radar input, each module consistently contributes to performance improvement. FLI provides reliable Gaussian initialization via depth-guided localization, IMA progressively refines multimodal aggregation, and MGF complements them with multi-level feature fusion. Together, these modules form a coherent framework that balances accuracy and efficiency while ensuring stable performance across different conditions.

\noindent\textbf{Ablation on FLI.}
As shown in Table \ref{tab:ablation_FLI}, FLI consists of three components: frustum-based initialization, integration of 4D radar, and depth back-projection. The initialization helps avoid resource wastage from out-of-view objects, while the 4D radar and depth back-projection provide prior knowledge, reducing the potential impact of unstable segmentation and depth estimation. Additionally, the coarse but structured Gaussian from FLI will be refined through IMA for eliminating instability.

\noindent\textbf{Ablation on IMA.}
IMA aggregates features from both image and radar modalities, and subsequently updates the gaussian positions within the frustum. Table \ref{tab:ablation_IMA} presents the contributions of these components. The integration of radar sparse pillars shows a relatively modest improvement, as feature aggregation has already been performed during the FLI phase and at the cross modal fuser.

\begin{table}[t]
    \belowrulesep=0pt
    \aboverulesep=0pt
    \centering
    \footnotesize
    \renewcommand\arraystretch{1.05}
    \begin{tabular}{c|c|c|c|c}
        \toprule[1.0pt]
        \multirow{2}{*}{\makebox[1.80cm]{Disturbance}} & \multicolumn{2}{c|}{$\text{mAP}_\text{EAA}$} & \multicolumn{2}{c}{$\text{mAP}_\text{DC}$} \\
        \cmidrule(lr){2-3} \cmidrule(lr){4-5}
        & \makebox[1.0cm]{LXL} & \makebox[1.0cm]{RaGS} & \makebox[1.0cm]{LXL} & \makebox[1.0cm]{RaGS} \\
        \midrule
        $\pm$ 0  $^\circ$, $\pm$ 0.0m & 56.42 & \textbf{61.86} & 73.03 & \textbf{81.63} \\
        $\pm$ 2  $^\circ$, $\pm$ 0.2m & 56.26 & \textbf{61.75} & 73.16 & \textbf{81.42} \\
        $\pm$ 5  $^\circ$, $\pm$ 0.5m & 50.25 & \textbf{56.66} & 72.07 & \textbf{76.39} \\
        % $\pm$ 10 $^\circ$, $\pm$ 1.0m & 39.47 & \textbf{45.55} & \textbf{61.29} & 58.23 \\
        \bottomrule[1.0pt]
    \end{tabular}
    \caption{Performance comparison under calibration disturbance.}
    \label{tab:calibration}
\end{table}

\begin{table}[t]
    \belowrulesep=0pt
    \aboverulesep=0pt
    \centering
    \footnotesize
    \renewcommand\arraystretch{1.05}
    \begin{tabular}{c|c|c|c|c}
        \toprule[1.0pt]
        \multirow{2}{*}{\makebox[1.80cm]{Condition}} & \multicolumn{2}{c|}{$\text{mAP}_\text{EAA}$} & \multicolumn{2}{c}{$\text{mAP}_\text{DC}$} \\
        \cmidrule(lr){2-3} \cmidrule(lr){4-5}
        & \makebox[1.0cm]{LXL} & \makebox[1.0cm]{RaGS} & \makebox[1.0cm]{LXL} & \makebox[1.0cm]{RaGS} \\
        \midrule
        Rain & 51.03 & \textbf{56.79} & 72.62 & \textbf{75.96} \\
        Haze & 51.16 & \textbf{56.72} & 69.49 & \textbf{72.29} \\
        Lowlight & 52.13 & \textbf{56.89} & 70.26 & \textbf{76.02} \\
        \bottomrule[1.0pt]
    \end{tabular}
    \caption{RaGS performance under simulated adverse weather.}
    \label{tab:weather}
\end{table}

% \begin{table}[t]
%     \belowrulesep=0pt
%     \aboverulesep=0pt
%     \centering
%     \footnotesize
%     \setlength{\tabcolsep}{11pt}
%     \renewcommand\arraystretch{1.05}
%     \begin{tabular}{c|c|c|c|c}
%         \toprule[1.0pt] 
%         \makebox[0.6cm]{$\mathcal{L}_{\text{det}}$} & \makebox[1.15cm]{w/ $\mathcal{L}_{\text{seg\_render}}$} & \makebox[1.15cm]{w/ $\mathcal{L}_{\text{depth\_render}}$} & \makebox[0.6cm]{$\text{mAP}_\text{EAA}$} & \makebox[0.6cm]{$\text{mAP}_\text{DC}$} \\
%         \midrule
%          \checkmark &  &  & 57.62 & 73.91 \\
%          \checkmark & \checkmark &  & 58.84 & 76.91 \\
%          \checkmark & & \checkmark & 58.99 & 76.63 \\
%          \checkmark & \checkmark & \checkmark & \textbf{59.45} & \textbf{76.98} \\
%         \bottomrule[1.0pt]
%     \end{tabular}
%     \caption{Ablation study on auxiliary loss of RaGS.}
%     \label{tab:ablation_aux}  % Keep this only once, after the caption.
% \end{table}

\begin{table}[t]
    \belowrulesep=0pt
    \aboverulesep=0pt
    \centering
    \footnotesize
    \renewcommand\arraystretch{1.05}
    \begin{tabular}{c|c|c|c|c|c}
        \toprule[1.0pt] 
        \makebox[1.75cm]{Models} & \makebox[0.8cm]{$\text{mAP}_\text{VoD}$} & \makebox[0.8cm]{$\text{mAP}_\text{TJ4D}$} &\makebox[0.8cm]{Param.} & \makebox[0.8cm]{FPS} & \makebox[0.8cm]{GFLOPs} \\
        \midrule
        RCFusion \cite{RCFusion} & 49.65 & 33.85 & 55.3M & 9.0 & 680.8 \\
        LXL \cite{LXL} & 56.31 & 36.32 & 64.0M & 6.1 & 1169.2\\
        HGSFusion \cite{HGSFusion} & 58.96 & 37.21 & 64.6M & 3.2 & 1899.6\\
        SGDet3D \cite{SGDet3D} & 59.75 & 41.82 & 73.6M & 9.2 & 1309.3\\
        \rowcolor{gray!20} RaGS (\textbf{ours}) & \textbf{61.86} & \textbf{41.95} & 56.5M & 10.5 & 781.0\\
        % \midrule
        % \multicolumn{6}{c}{Analysis of RaGS}\\
        % \midrule
        % Metrics & Extra. & FLI & IMA & MGF & ALL \\
        % \midrule
        % Param. (M) & 38.4 & 1.9 & 5.9 & 10.3 & 56.5\\
        % FPS & 48.3 & 376.8 & 15.9 & 109.1 & 10.5\\
        % GFLOPS & 508.5 & 4.2 & 162.8 & 98.3 & 781.0\\
        \bottomrule[1.0pt]
    \end{tabular}
    \caption{Comparison with state-of-the-art models.}
    \label{tab:ablation_comparison}
\end{table}

\begin{table}[t]
    \belowrulesep=0pt
    \aboverulesep=0pt
    \centering
    \footnotesize
    \renewcommand\arraystretch{1.05}
    \begin{tabular}{c|c|c|c|c|c}
        \toprule[1.0pt] 
        \makebox[1.75cm]{Metrics} & \makebox[0.8cm]{Extra.} & \makebox[0.8cm]{FLI} &\makebox[0.8cm]{IMA} & \makebox[0.8cm]{MGF} & \makebox[0.8cm]{All} \\
        \midrule
        Param. (M) & 38.4 & 1.9 & 5.9 & 10.3 & 56.5\\
        FPS & 48.3 & 376.8 & 15.9 & 109.1 & 10.5\\
        GFLOPS & 508.5 & 4.2 & 162.8 & 98.3 & 781.0\\
        \bottomrule[1.0pt]
    \end{tabular}
    \caption{Analysis of efficiency across RaGS components.}
    \label{tab:RaGS_details}
\end{table}

\noindent\textbf{Ablation on Hyper-parameters.}
We conduct an ablation study on the the anchor number \(N\), as shown in Table \ref{tab:ablation_num}, where both mAP and GFLOPS are reported. To balance performance and computational cost, we select 2-level MGF fusion, 3 layers for IMA, and 12,800 Gaussian anchors. For fusion level \(L\) in MGF and the aggregation number \(M\) in IMA, see supplementary material for further hyperparameter ablation details. 
% Additionally, Table S2 in the supplementary material shows the contribution of auxiliary tasks to 3D object detection performance, which we offer detailed analysis.

\subsection{Robustness Analysis}
As shown in Table \ref{tab:calibration} and Table \ref{tab:weather}, RaGS maintains stable performance under calibration disturbance and adverse weather. Even with $\pm$5$^\circ$/$\pm$0.5m perturbations or simulated challenging weather condition (rain, haze, low light), both mAP$_\text{EAA}$ and mAP$_\text{DC}$ remain consistently higher than LXL \cite{LXL}. This demonstrates that the proposed frustum-guided localization and Gaussian fusion ensure robust and reliable detection across diverse real-world conditions.

\subsection{Computational Cost}
% \noindent\textbf{Comparison with State-of-the-art Models.}
We compared RaGS with existing 4D radar and camera fusion models on mAP, FPS, and GFLOPs. As shown in Table \ref{tab:ablation_comparison}, RaGS outperforms with fewer parameters and lower GFLOPS. Unlike BEV-grid methods, RaGS uses iterative refinement to focus on sparse objects while maintaining scene understanding. SGDet3D uses 160$\times$160$\times$8 fixed sampling points, about 16 times more than our 12,800. Table \ref{tab:RaGS_details} presents efficiency analysis of each component within the RaGS Framework.

\section{Conclusions}
In this work, we propose RaGS, the first framework to leverage 3D Gaussian Splatting for fusing 4D radar and monocular images in 3D object detection. RaGS models the fused multi-modal features as continuous 3D Gaussians, enabling dynamically object attention while preserving comprehensive scene perception. The architecture, consisting of FLI, IMA, and MGF, progressively constructs and refines the Gaussian field through frustum-based localization, cross-modal aggregation, and multi-level fusion. Experiments on public benchmarks show state-of-the-art performance. 

\emph{Limitations.} Our ongoing work focuses on extending it to sequential modeling by explicitly exploiting radar velocity within 4D GS for enhanced temporal object awareness.

{
    \small
    \bibliographystyle{cvpr2026}
    \bibliography{cvpr2026}
}


\end{document}